# Transportation Density Reduction Caused by City Lockdowns Across the World during the COVID-19 Epidemic: From the View of High-resolution Remote Sensing Imagery

Chen Wu, *Member,* Sihan Zhu, Jiaqi Yang, Meiqi Hu, Bo Du, *Senior Member, IEEE*, Liangpei Zhang, *Fellow*, *IEEE,* Lefei Zhang, *Senior Member, IEEE,* Chengxi Han, and Meng Lan

*Abstract*—As the COVID-19 epidemic began to worsen in the first months of 2020, stringent lockdown policies were implemented in numerous cities throughout the world to control human transmission and mitigate its spread. Although transportation density reduction inside the city was felt subjectively, there has thus far been no objective and quantitative study of its variation to reflect the intracity population flows and their corresponding relationship with lockdown policy stringency from the view of remote sensing images with the high resolution under 1m. Accordingly, we here provide a quantitative investigation of the transportation density reduction before and after lockdown was implemented in six epicenter cities (Wuhan, Milan, Madrid, Paris, New York, and London) around the world during the COVID-19 epidemic, which is accomplished by extracting vehicles from the multi-temporal high-resolution remote sensing images. A novel vehicle detection model combining unsupervised vehicle candidate extraction and deep learning identification was specifically proposed for the images with the resolution of 0.5m. Our results indicate that transportation densities were reduced by an average of approximately 50% (and as much as 75.96%) in these six cities following lockdown. The influences on transportation density reduction rates are also highly correlated with policy stringency, with an $R^2$ value exceeding 0.83. Even within a specific city, the transportation density changes differed and tended to be distributed in accordance with the city's land-use patterns. Considering that public transportation was mostly reduced or even forbidden, our results indicate that city lockdown policies are effective at limiting human transmission within cities.

This work was supported in part by the National Natural Science Foundation of China under Grant 61971317 and 41801285. (Corresponding author: Bo Du, Liangpei Zhang)

C. Wu, S. Zhu, J. Yang, M. Hu, L. Zhang and C. Han are with the State Key Laboratory of Information Engineering in Surveying, Mapping and Remote Sensing, Wuhan University, Wuhan 430072, China (e-mail: chen.wu@whu.edu.cn; zhusihan96@whu.edu.cn; jiaqiy@whu.edu.cn; meiqi.hu@whu.edu.cn; zlp62@whu.edu.cn; chengxihan@whu.edu.cn ).

B. Du, L. Zhang, and M. Lan is with the School of Computer Science, Wuhan University, Wuhan 430072, China (email: gunspace@163.com; zhanglefei@whu.edu.cn; menglan@whu.edu.cn ).

## I. INTRODUCTION

In 2020, the spread of the novel coronavirus (COVID-19) completely changed human life across the whole world [1, 2].

In order to control the COVID-19 epidemic, the city of Wuhan firstly implemented a stringent lockdown policy [3, 4], which appears to be the largest quarantine to date and was unprecedented in human public health history [5, 6]. As the global COVID-19 epidemic worsened, many other cities around the world subsequently implemented similar lockdown policies, including limiting public transportation, closing schools and non-essential shops, ordering residents to stay and work at home, prohibiting public events and gatherings, etc., to control transmission [7-14]. Numerous studies have also indicated that such control measures can mitigate the growth of the COVID-19 epidemic [6, 15-20].

Since the details of lockdown policies and their degrees of stringency have varied across different cities [21], it is difficult to quantitatively evaluate their effects and draw fair global comparisons. Transportation density is a good indicator of intracity population flow, as reductions in this metric have certainly been felt by the inhabitants of the cities under lockdown. However, it is difficult to obtain accurate and quantitative transportation data from various cities in different countries.

High-resolution remote sensing data can provide objective and consistent observations among different cities and between two periods (before and after city lockdown) [22-25], which offers a unique opportunity to evaluate the transportation density reduction caused by city lockdowns during the COVID-19 epidemic for a fair comparison [26-28]. Although a recent work addressed an analysis of traffic pattern during COVID-19 epidemic, it used the images with the resolution of 3m, which is hard to identify normal vehicles accurately with such a resolution [29]. For most existing studies focusing on vehicle detection on remote sensing imagery, they are based on the aerial images with limited experimental areas, since their resolutions are smaller than 0.3m and have enough spatial details to identify cars [30-37]. The problem is, only a few satellite sensors can reach the resolution of 0.3m, and they are not just in the right pathway to acquire the necessary images for studying city lockdown. Hence, it is also necessary to develop a robust vehicle extraction model for remote sensing images with a lower and more common resolution, which can be employed in the complex scenes of practical applications covering the whole city.

**Table 1:** Metadata of the collected high-resolution remote sensing images covering the early epicenters of COVID-19 epidemic, including Wuhan, Milan, Madrid, Paris, London, and New York. These images were all acquired both before and after city lockdown with the smallest intervals possible. Some images are mosaics of two or three images, with their acquisition data and time shown in the bracket.

| City | Lockdown Date | Image Date | Image Time (Local) | Weekday/ Weekend | Image Area ($km^2$) | Resolution (m) |
|---|---|---|---|---|---|---|
| **Wuhan** | 2020/1/23 [*1] | 2019/10/17 | 11:26:26 | Thursday | 171.0 | 0.50 |
| | | 2020/2/4 | 11:09:42.1 (+10s) | Tuesday | | 0.58 |
| **Milan** | 2020/3/8 [*2] | 2019/10/16 | 11:33:57.3 | Wednesday | 32.6 | 0.51 |
| | | 2020/3/28 | 11:22:47.0 | Saturday | | 0.58 |
| **Madrid** | 2020/3/14 [*3] | 2019/7/18 | 12:17:06.8 (+18.7s) | Thursday | 65.8 | 0.50 |
| | | 2020/4/3 | 12:16:33.8 | Friday | | 0.50 |
| **Paris** | 2020/3/17 [*4] | 2020/1/18 | 11:59:30.5 | Saturday | 89.0 | 0.53 |
| | | 2020/3/28 | 12:10:46.3 | Saturday | | 0.64 |
| **New York** | 2020/3/22 [*5] | 2020/2/19 | 11:01:03.1 (+23.5s) | Wednesday | 70.0 | 0.52 |
| | | 2020/4/6 | 10:49:46.1 (+30.4s/+50.9s) | Monday | | 0.53 |
| **London** | 2020/3/23 [*6] | 2019/5/14 | 11:14:19.5 | Tuesday | 46.6 | 0.62 |
| | | 2020/4/21 2020/4/23 | 11:09:17.5 (+15m 24.1s) | Monday/ Wednesday | | 0.54 |

*1. Wuhan municipal headquarters for the COVID-19 epidemic prevention and control. 2020, Announcement No. 1. http://www.gov.cn/xinwen/2020-01/23/content_5471751.htm.

*2. Presidenza del Consiglio dei Ministri. 2020, Coronavirus, le misure adottate dal Governo. http://www.governo.it/coronavirus-misure-del-governo.

*3. El País. 2020, Lack of testing hampering Spain's efforts to slow coronavirus outbreak. https://english.elpais.com/society/2020-03-18/lack-of-testing-hampering-spains-efforts-to-slow-coronavirus-outbreak.html.

*4. The Independent. 2020, Coronavirus: France imposes 15-day lockdown and mobilises 100,000 police to enforce restrictions. https://www.independent.co.uk/news/world/europe/coronavirus-france-lockdown-cases-update-covid-19-macron-a9405136.html.

*5. New York State. 2020, Governor Cuomo Signs the 'New York State on PAUSE' Executive Order. https://www.governor.ny.gov/news/governor-cuomo-signs-new-york-state-pause-executive-order.

*6. Prime Minister's Office 10 Downing Street, & The Rt Hon Boris Johnson MP. 2020, Prime Minister's statement on coronavirus (COVID-19): 23 March 2020 https://www.gov.uk/government/speeches/pm-address-to-the-nation-on-coronavirus-23-march-2020.

In the present study, we use the number of vehicles extracted from multi-temporal high-resolution remote sensing (RS) images before and after lockdown in six cities around the world, with the goal of conducting an objective and quantitative study of transportation density reduction caused by city lockdown during the COVID-19 epidemic. A novel vehicle detection model that combines a morphology filter and multi-branch deep learning identification network was developed to quantify the impact of city lockdown on transportation density. We further aim to analyze the correlation between transportation density reduction and lockdown stringency, as well as to provide an interpretation of the geographic view within the cities.

We choose six large and famous cities for analysis: Wuhan, Milan, Madrid, Paris, New York, and London, all of which were epicenters in the early COVID-19 epidemic [38] and subsequently implemented stringent lockdown measures [4, 7, 9, 10, 13, 14]. Multi-temporal remote sensing images with resolution of 0.5m covering the core areas of these cities, acquired before and after the implementation of city lockdown policy, were collected for a total study area of $475 km^2$ at one time. Open Street Map was used to extract road regions [39], while a COVID stringency index [21] was utilized to evaluate the city lockdown policy.

References for a total area of $63 km^2$ and 35770 labeled vehicles indicated that our vehicle detection results obtain a total accuracy of 69.63%, which is satisfactory at this resolution in practical applications. Following analysis of the vehicle density changes, our results indicate that the transportation densities reduced by an average of approximately 50% in these six cities after city lockdown, with a decline sometimes as high as 75.96%. The influence on transportation density reduction rates is highly correlated with policy stringency, with an $R^2$ exceeding 0.83. In terms of the view inside the city, transportation density changes also varied according to city land-use patterns. Considering that public transport was either limited or forbidden during these periods [4, 40-42], city lockdown policy can be deemed effective in controlling intracity human transmission.

## II. STUDY SITE AND REMOTE SENSING IMAGES

In order to objectively compare the transportation density variations before and after lockdown in the six large cities selected, we collected multi-temporal high-resolution RS images to extract the numbers of vehicles on the road. The metadata of all RS images used in this study is presented in Table 1. Almost all RS images were acquired by the Pleiades satellite, with a resolution of nearly 0.5 m and four spectral bands (Blue, Green, Red, and Near-Infrared). The only exception is the RS image covering Wuhan before lockdown. The nearest available and clear Pleiades image covering the study area with low cloud coverage was acquired in 2016, which is too long ago compared with the other image pairs. Accordingly, we selected the RS images from WorldView-3 instead, then resampled them to a resolution of 0.5 m and four spectral bands.

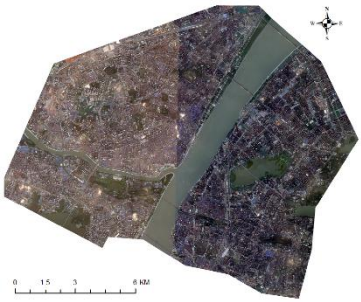 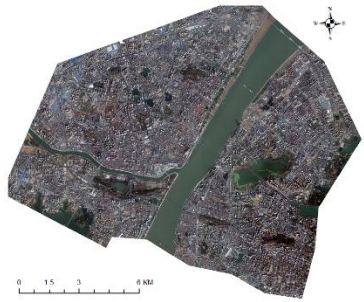 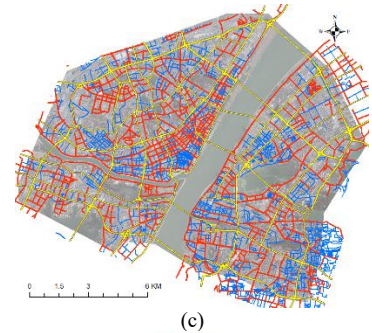

(a)        (b)        (c)

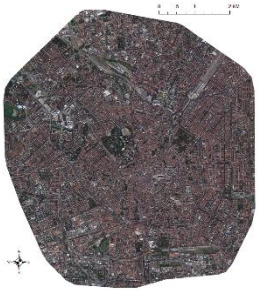 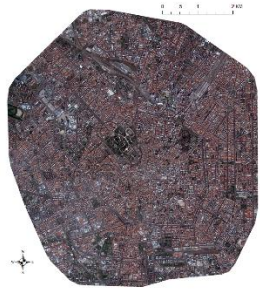 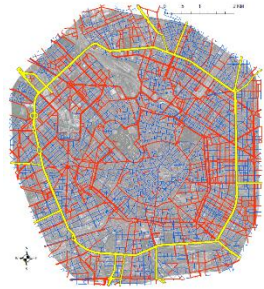

(d)        (e)        (f)

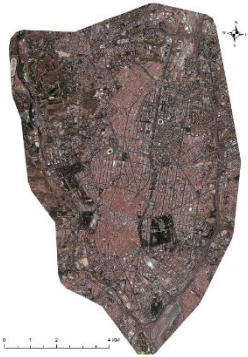 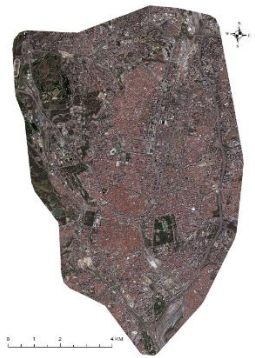 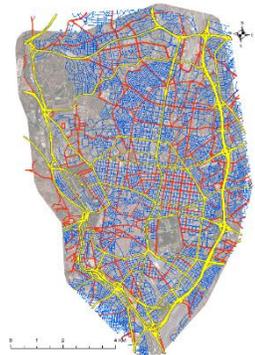

(g)        (h)        (i)

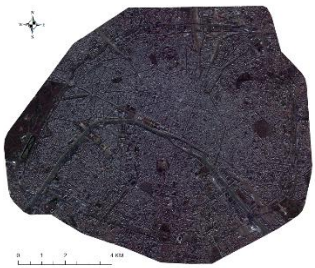 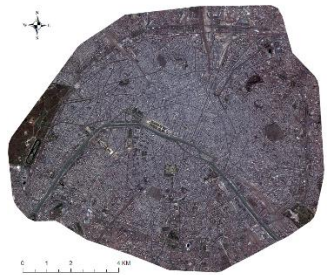 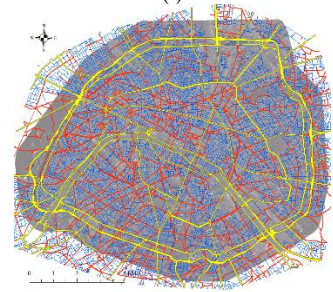

(j)        (k)        (l)

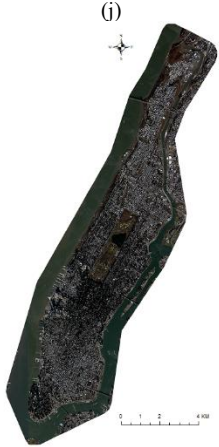 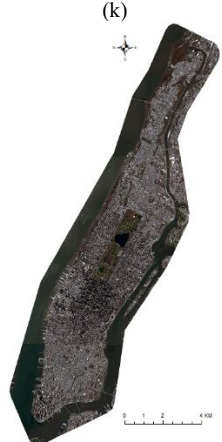 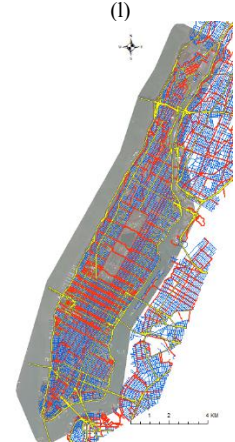

(m)        (n)        (o)

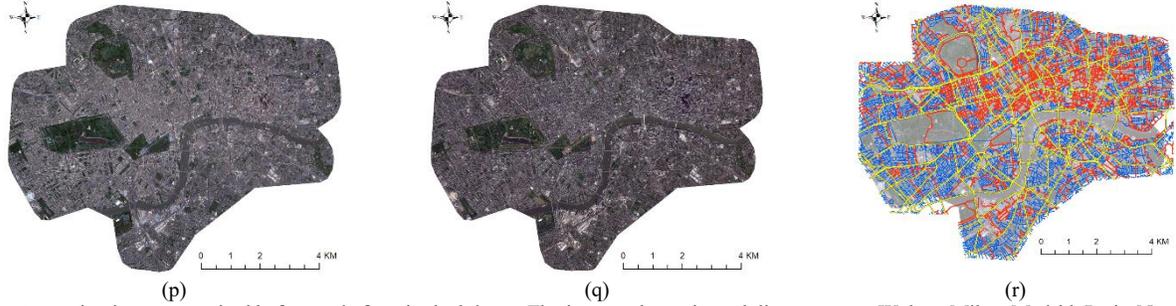

Figure 1: Remote sensing images acquired before and after city lockdown. The images shown in each line represent Wuhan, Milan, Madrid, Paris, New York, and London respectively. The first column presents the images before city lockdown, while the second column shows the images after city lockdown. Moreover, the third column shows the road networks from OSM; here, yellow represents arterial roads, red represents collector roads and blue represents local roads.

The city lockdown dates and subsequent remote sensing image acquisition time are presented alongside the cumulative confirmed cases [43-48] and COVID-19 government response stringency index developed by Oxford University [21] in Figure 2. It can be observed that the RS images were all acquired in the rising phases of the COVID-19 epidemic when the governments were implementing the most stringent lockdown measures. These collected remote sensing data will therefore be effective for evaluating the influence of lockdown policy on transportation density.

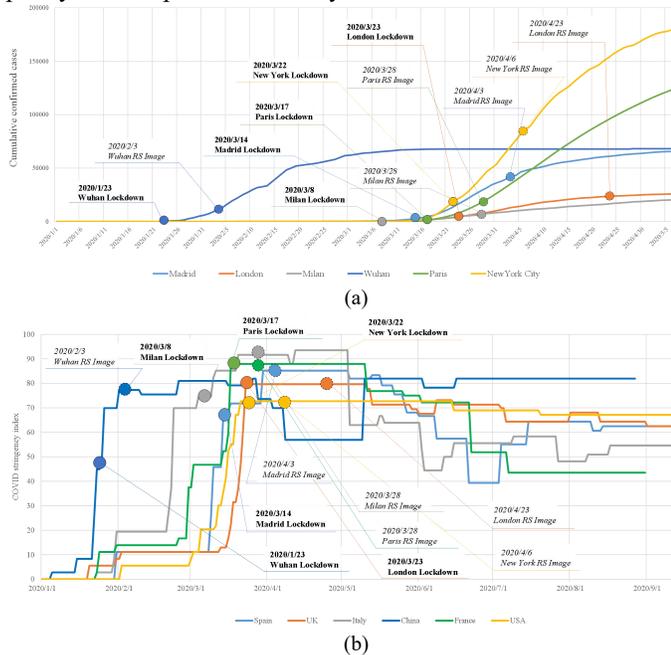

Figure 2: City lockdown and subsequent remote sensing image acquisition date are shown corresponding to the cumulative confirmed cases and COVID stringency index.

Since these six cities differ markedly in terms of size and area, we only selected their core regions with obvious landmarks. (1) For the city of Wuhan, the study area covers the region inside the second ring road; (2) For the city of Milan, the study area was bounded by SS11 road; (3) For the city of Madrid, we selected the region contained by the M-30 road as the study site; (4) For the city of Paris, the study area contains all arrondissements except for two parks; (5) For New York City, the study area covers the borough of Manhattan; (6) For the city of London, we selected the region covered by zone 1. The study areas and multi-temporal high-resolution remote sensing images are presented in Figure 1. The total observation area at one time is $475km^2$.

In order to identify vehicles on the roads only, we selected Open Street Map (OSM) data for the six cities. The OSM data for "motorway", "trunk", "primary" and their corresponding "link" were classified as arterial roads with a buffer of 20 m, while those for "secondary", "tertiary", "unclassified" and their corresponding "link" were classified as collector roads with a buffer of 20 m; moreover, those for "residential" and "living_street" were classified as local roads with a buffer of 15 m [49]. Only the vehicles within these buffers were extracted and counted. The OSM data are presented in Figure 1. All of these images were processed using the GS pan-sharpening method and co-registered to the OSM maps to facilitate better alignment.

## III. METHODOLOGY

### A. Vehicle Candidate Extraction

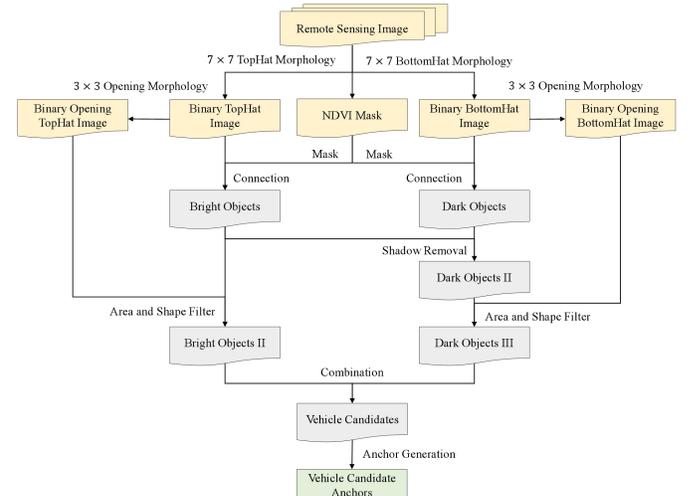

Figure 3: Flowchart of Vehicle Candidate Extraction

Since the vehicles identified in the remote sensing images with resolution of 0.5m cannot provide enough shape information to facilitate direct detection, we decide to combine unsupervised vehicle candidate extraction with deep learning identification in our model. The goal of the unsupervised vehicle candidate extraction is to find possible vehicle targets and remove interference on the road, while the deep learning identification is used to distinguish between vehicles and non-vehicles.

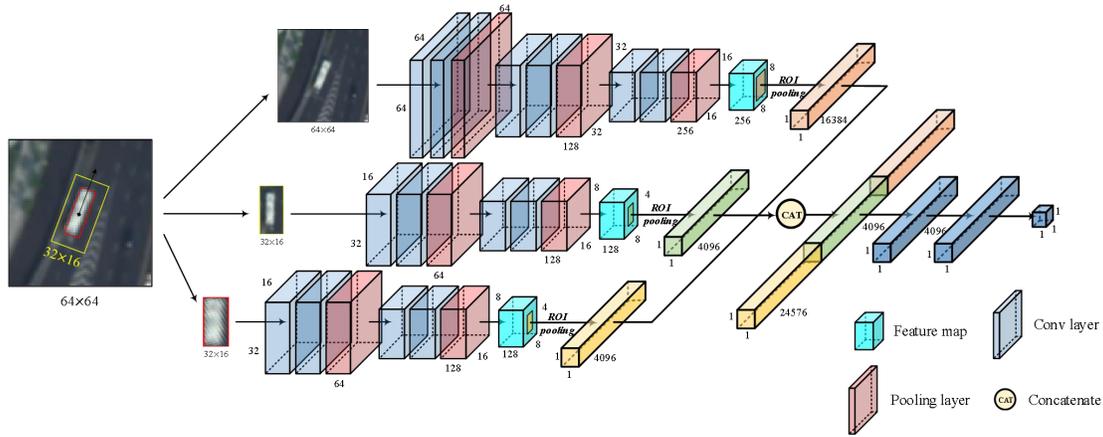

Figure 4: Diagram of multi-branch deep learning model

During the process of vehicle candidate extraction, the basic concept is that vehicles in remote sensing images contrast with the road background, enabling them to be captured by human interpreters. Accordingly, we first separate the bright and dark objects by segmenting the contrast density images using the TopHat and BottomHat morphology process [50]; subsequently, the connected objects are filtered according to their areas, shapes, and compactness, as shown in Figure 3. In more detail, the procedure is as follows:

In this process, only the road regions masked by OSM data are considered. $7 \times 7$ TopHat and BottomHat morphology processes are utilized for each spectral band of the remote sensing image to highlight bright and dark objects against the road background. Two contrast density images are generated by the Euclidean distance of the multi-band morphology results, after which two binary images are obtained by separately thresholding TopHat and BottomHat density images in accordance with visual interpretation. The corresponding opening binary images are produced by means of a $3 \times 3$ opening morphology process on the binary TopHat and BottomHat images and used as auxiliary information. Here, the opening process is used to avoid mis-detection caused by the connection of multiple objects. Moreover, a vegetation mask is generated using the manual thresholding of the NDVI feature image to avoid false detection of tree shadows. We now have five types of input data: binary TopHat image, binary BottomHat image, binary Opening TopHat image, binary Opening BottomHat image, and NDVI mask.

In the first step, bright and dark objects are extracted separately by connecting the binary TopHat image and binary BottomHat image with the vegetation mask.

For dark objects, shadows always accompany bright vehicles and can thus easily be falsely detected as dark vehicles. Thus, dark objects are taken out by means of shadow removal if a certain proportion (e.g. 30%) of the contours of a dark object are adjacent to bright objects by one or two pixels. Notably, dark objects with fairly good vehicle shapes are retained.

In the next step, bright objects and dark objects are filtered according to their areas, shapes and compactness. There are three criteria: 1) the area of objects should be neither too large or too small; 2) the height and width of the minimum bounding rectangle should be smaller than the thresholds, while the aspect ratio should also not be too large; 3) the proportions of an object occupying its contour hull or minimum bounding box should be larger than a given threshold. Moreover, objects filtered by the third criterion may be falsely removed due to the connection of two nearby vehicles in the image with such a resolution. Therefore, for these removed objects, their corresponding objects in the binary opening image will be added instead, provided that the alternative opening objects satisfy the stricter criterion discussed above.

Finally, the bright objects and dark objects are combined to make up the vehicle candidates. However, since the pixels in these candidates may not be especially accurate, while some vehicles may be divided into two objects for a number of reasons (e.g. a dark sunroof), we generate anchors that center the vehicle candidates. The bounding boxes of vehicle candidates are expanded with a height zoom ratio of [1, 1.5, 2] and width zoom ratio of [1, 1.25, 1.5], after which the anchor ratios are rotated by 90°. Normally, for each vehicle candidate, we will obtain 18 anchors; however, if the vehicle candidates have an obvious direction with a large aspect ratio, the anchor expansion will only be generated in this direction.

### B. Deep Learning Identification

After the vehicle anchors are obtained, we develop a multi-branch CNN model to identify and distinguish between vehicles and non-vehicles. Since vehicles on remote sensing images with such a resolution do not contain detailed shape information, we fuse multi-scale patches with R, G, B and NIR bands that center the anchors to be the input of the deep learning model. The first input is a window patch of size $64 \times 64$, which is clearly larger than the vehicle, to facilitate extracting the background information; the second input is a sub-window patch of size of $32 \times 16$, aligned with the direction of the anchor; moreover, the third input is the aligned anchor patch, which is resized to $32 \times 16$ for convenience.

The multi-branch deep learning model is shown in Figure 4. The structure of this model derives from VGG-16 [51]; it contains convolution layers, max pooling layers, ROI pooling layers, and fully connected layers. ROI pooling layers are used here to guarantee a fixed size for the output vector in each branch. The final layer uses a sigmoid function to output the probability that the detected object is a vehicle. When training the deep learning model, we train each branch separately by connecting the features after ROI pooling to three new fully connected layers, as in the multi-branch model. The three trained branches are then connected, as in Figure 4, with new fully connected layers, then re-trained with the same training samples. Since we have extracted numerous vehicle candidates, it is much easier to label the training samples from the candidates rather than drawing boxes directly from the images. The number of training samples selected in each image is presented in Table 2. All training and testing samples are normalized with reference to the statistics of their source images. Random flipping (both vertically and horizontally) is adopted as a data augmentation method during the training process.

**Table 2: Training samples for vehicles and non-vehicles in high-resolution remote sensing images**

| City | Image Date | Vehicle Samples | Non-Vehicle Samples |
|------|-----------|-----------------|---------------------|
| Wuhan | 20191017 | 965 | 2985 |
| | 20200204 | 479 | 2969 |
| Milan | 20191016 | 700 | 2800 |
| | 20200328 | 700 | 2800 |
| Madrid | 20190718 | 758 | 2890 |
| | 20200403 | 913 | 3502 |
| Paris | 20200118 | 987 | 3263 |
| | 20200328 | 949 | 3335 |
| New York | 20200219 | 1076 | 2851 |
| | 20200406 | 1006 | 2910 |
| London | 20190514 | 700 | 2800 |
| | 20200421 | 700 | 2800 |
| **Total** | | **9933** | **35905** |

In the optimization procedure, binary cross-entropy is used as the loss function. Since a quantity imbalance exists between the vehicle and non-vehicle training samples, we use a simple weighting approach in the loss function, which assigns $\frac{n_v+n_n}{2n_v}$ to vehicle samples and $\frac{n_v+n_n}{2n_n}$ to nonvehicle samples; here, $n_v$ and $n_n$ are the sample amounts of vehicles and non-vehicles. This simple approach is able to balance the weights of vehicle and non-vehicle samples during the calculation of loss. The optimizations of each single branch and the multi-branch model are all implemented 100 times, and the Adam optimizer is applied [52]; the batch size is set to 200, and the shuffle process is also used. More specifically, we use a warmup learning rate scheduler strategy [53], where the start rate is $1e^{-4}$, the max rate is $1e^{-3}$, the min rate is $1e^{-6}$, the warmup epoch is 20, the sustaining epoch is 0, and the decaying parameter is 0.8. For the training of the first branch with window patch, the start rate and max rate are changed to be $1e^{-5}$ and $1e^{-4}$.

In the testing procedure, the testing samples are determined to be vehicles if the output probability of the deep learning model exceeds 0.5.

*C. Post-processing*

In the post-processing phase, there are two steps: the first is Non-Maximum Suppression (NMS), while the second is the removal of vehicles under temporal shadow coverage.

Since numerous anchors have been generated that center the vehicle candidates, several of them will be retained after deep learning refining is complete. Therefore, we utilize NMS to choose the most probable detection box. Unlike the normal NMS, we apply some special processes here. First, if the heights, widths, and aspect ratios of the anchors do not satisfy the shape filter criterion, they will be suppressed. Second, in addition to the Intersection over Union (IoU), the Intersection over Area (IoA) of the anchor with a maximum probability is also used to filter repetitive anchors. Third, if a suppressed anchor shows a probability that is only 0.05 less than the remaining one, but has the minimum area, it will be retained instead. After the NMS process is complete, the detection results for vehicles on the high-resolution remote sensing images are obtained.

However, since some of these images were acquired in winter, the solar altitudes are still low even when the acquisition time is midday. This phenomenon leads to more shadow coverage appearing in only one image of the multi-temporal pair, which will result in unfair comparison of vehicle numbers on the road. We therefore utilize a Tasi's ratio image to find and extract shadow coverages [54]. Shadow areas are segmented by means of manual thresholding. A $5 \times 5$ Closing morphology process and $9 \times 9$ Opening morphology process are subsequently implemented so that only large-area shadow regions remain. The union of shadow areas in a multi-temporal image pair is used to remove the covering vehicles in both images. All results and statistics presented in this paper are based on the detection maps following multi-temporal shadow coverage removal.

## IV. EXPERIMENTS AND RESULTS

*A. Results and Accuracy Evaluation*

The examples of vehicle detection results are shown in Figure 5. It can be observed that after city lockdown, the vehicle densities in these images obviously reduced. Most vehicles in these images were successfully detected, and the vehicle numbers were accurately estimated.

In order to demonstrate the credibility of our method, we selected nine regions for each RS image (for a total of 108 reference regions) and manually labelled all vehicles. The total size of the reference area extends to $63km^2$ with 35770 labeled vehicles. Since the objective of this study is to count the number of vehicles, a detection will be judged to be true if the detection box covers the labeled box.

The total accuracy of the proposed model, comparison methods and the corresponding ablation experiments are presented in Table 3; here, "All Image Training" represents the proposed model. The comparison method selected is the well-known faster RCNN with rotating box. Since the training sample set for our proposed model is not suitable for faster RCNN, we attempt two ways of implementing this model. The first way involves downsampling the DOTA dataset to a resolution similar to that of our data, then training the faster RCNN model for vehicle detection [55]. In the second way, we conduct fine-tuning with the pre-trained model, as in the first way, based on half of the references. As Table 3 indicates, the classical faster RCNN obtains far lower accuracies than the proposed model; this is because the spatial shape information is lacking in images with a resolution of 0.5m, which are unsuitable for direct object detection.

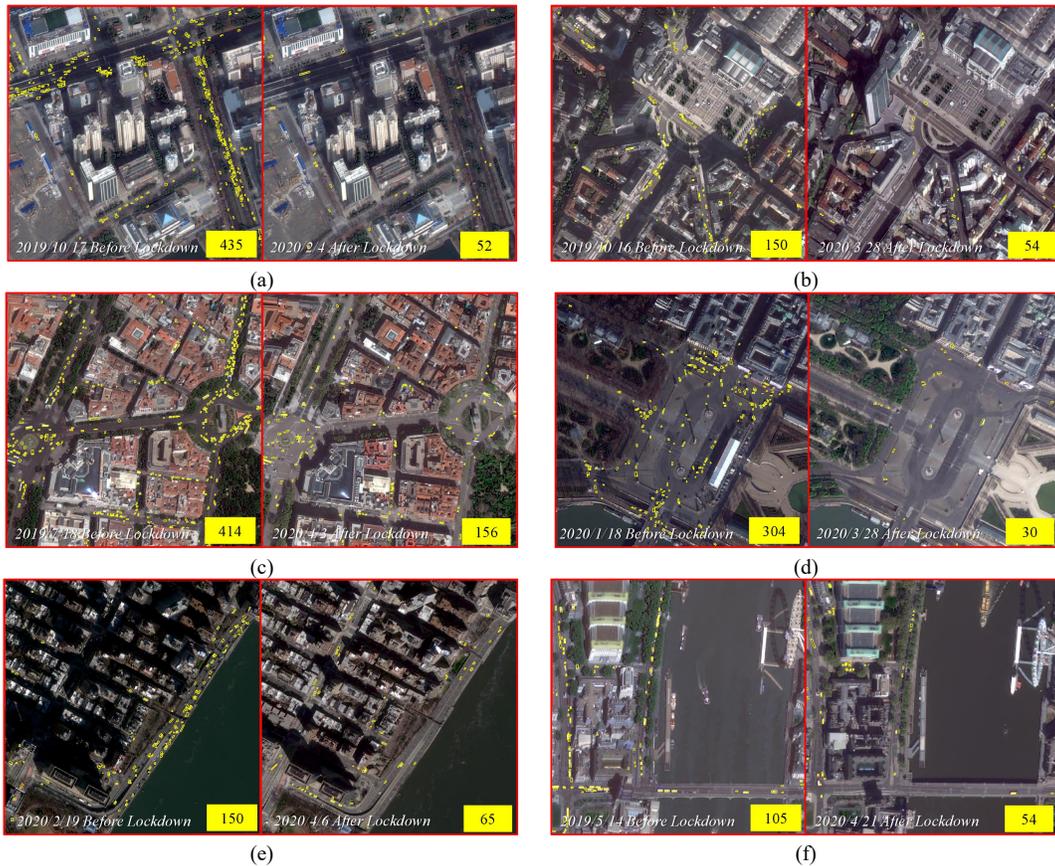

Figure 5: Vehicle detection results in multi-temporal high-resolution remote sensing images, labelled with yellow boxes. (a) Zoomed-in images covering the road near Hankou Railway Station in Wuhan; (b) Zoomed-in images covering the Centrale FS in Milan; (c) Zoomed-in images covering the Puerta de Alcala in Madrid; (d) Zoomed-in images covering the Place de la Concorde in Paris; (e) Zoomed-in images covering the road near the United Nations Headquarters in New York City; (f) Zoomed-in images covering the Westminster Bridge in London. Vehicles in regions covered by shadows within any multi-temporal RS images have been removed to facilitate fair comparison. The numbers in yellow rectangles indicates the number of vehicles detected in the image by our model.

**Table 3:** Total accuracy of the proposed model, comparison methods, and ablation experiments.

|  | Precision Rate | Recall Rate | F1 Score |
|---|---|---|---|
| Vehicle Candidate Extraction | 11.66% | **85.42%** | 20.32% |
| Single Image Training | 64.80% | 65.14% | 64.60% |
| All Images Training with Single Image Fine-tune | 75.77% | 62.27% | 68.08% |
| All Images Training without Weighting | 68.83% | 69.61% | 68.99% |
| All Images Training without Anchor Generation | **79.19%** | 57.60% | 66.39% |
| All Images Training only for Anchor Branch | 33.57% | 80.06% | 46.63% |
| All Images Training only for Sub-Window Branch | 71.94% | 64.71% | 67.65% |
| All Images Training only for Window Branch | 69.04% | 64.69% | 66.41% |
| All Images Training for Sub-Window + Anchor Branch | 65.45% | 70.12% | 67.44% |
| All Images Training for Window + Anchor Branch | 61.36% | 72.93% | 66.35% |
| All Images Training for Window + Sub-Window Branch | 74.90% | 63.23% | 68.25% |
| Faster RCNN Trained with DOTA | 30.36% | 9.32% | 10.36% |
| Faster RCNN Fine-tune with Reference | 42.36% | 6.68% | 11.03% |
| All Images Training | 69.86% | 69.90% | **69.63%** |

As shown in Table 3, vehicle candidate extraction obtains a recall rate of 85.42%, meaning that it has found most vehicles in the image. The objective of deep learning identification is to increase the precision rate while simultaneously maintaining the recall rate. After DL refining is complete, it can be observed that the precision rate increases very substantially (11.66% to 69.86%), while the recall rate is kept relatively steady (85.42% to 69.90%). Single Image Training means that the DL model for each image is trained only with the training samples themselves. The accuracy results show that more training samples can yield more robust performance, where the best model is trained using all training samples. Moreover, we also test the result of the DL model that is trained with all samples and fine-tuned specific to the testing image. We find that the fine-tuned model identifies vehicles more accurately, but also misses more targets; this may be because the fine-tuning increases the model's identification ability, but will also falsely remove some vehicles if they are not covered in the training samples of the testing image.

**Table 4: Statistics of vehicle numbers in the six cities before after lockdown, along with their change ratios.**

| City | Acquisition Time | Total Vehicle Number | Vehicle Number on Arterial Road |
|------|-----------------|---------------------|-------------------------------|
| **Wuhan** | Before Lockdown | 59338 | 25736 |
| | After Lockdown | 24620 | 6186 |
| | *Change Ratio* | ***-58.51%*** | ***-75.96%*** |
| **Milan** | Before Lockdown | 35908 | 4316 |
| | After Lockdown | 20929 | 1596 |
| | *Change Ratio* | *-41.71%* | *-63.02%* |
| **Madrid** | Before Lockdown | 64581 | 18376 |
| | After Lockdown | 66327 | 9649 |
| | *Change Ratio* | *+2.70%* | *-47.49%* |
| **Paris** | Before Lockdown | 35449 | 16964 |
| | After Lockdown | 19326 | 6000 |
| | *Change Ratio* | *-45.48%* | *-64.63%* |
| **New York** | Before Lockdown | 33408 | 15358 |
| | After Lockdown | 38101 | 12632 |
| | *Change Ratio* | *+14.05%* | *-17.75%* |
| **London** | Before Lockdown | 33782 | 9231 |
| | After Lockdown | 37004 | 6585 |
| | *Change Ratio* | *+9.54%* | *-28.66%* |

The proposed weighting and anchor generation processes are also evaluated by ablation. The experiments show that anchor generation increases the total accuracy (F1 score) by 3%, while the weighting process increases the accuracy by less than 1%; since the imbalance between vehicle and non-vehicle samples is not severe, the weighting process improves the performance only slightly. Moreover, as the proposed model consists of three branches, we test various combinations of different branches. The ablation experiments reveal that the proposed method with all three branches outperforms those with only one or two branches.

**Table 5:** Vehicle detection accuracy in all images covering six cities

| City | Image Date | Precision Rate | Recall Rate | F1 Score |
|------|-----------|----------------|-------------|----------|
| **Wuhan** | 20191017 | 76.28% | 66.85% | 71.26% |
| | 20200204 | 63.32% | 65.91% | 64.59% |
| **Milan** | 20191016 | 70.20% | 82.06% | 75.67% |
| | 20200328 | 74.90% | 67.76% | 71.15% |
| **Madrid** | 20190718 | 84.61% | 71.36% | 77.42% |
| | 20200403 | 74.52% | 69.43% | 71.89% |
| **Paris** | 20200118 | 62.63% | 65.91% | 64.23% |
| | 20200328 | 62.17% | 55.85% | 58.84% |
| **New York** | 20200219 | 64.13% | 75.46% | 69.34% |
| | 20200406 | 71.17% | 65.51% | 68.22% |
| **London** | 20190514 | 66.44% | 75.60% | 70.72% |
| | 20200421 | 67.90% | 77.06% | 72.19% |
| **Total** | | 69.86% | 69.90% | 69.63% |

The detailed accuracies for the six cities are listed in Table 5 It can be observed from the table that almost all results have accuracies higher than 65%. These accuracies are all satisfactory in practical applications, especially considering the lack of detailed shape information, the complex acquisition environments, and the large covering areas.

### B. Changes of Transportation Density

The vehicle number statistics before and after lockdown, along with their change ratios, are presented in Table 4. It can be observed that in the entire study area, the cities of Wuhan, Milan and Paris exhibited obvious reductions, at the rate of approximately -50%, while the other three cities showed a slight increase.

It is worth noting that, in these cities, stopped vehicles may be parked along the roadside, meaning that it is difficult to distinguish between parked vehicles and running vehicles on the same road. Therefore, in order to better evaluate the real influence of city lockdown on transportation, we create statistics for vehicles on the arterial roads in each city, which are shown in the second column of Table 4. As the table indicates, city lockdown obviously decreases the vehicle densities in the main traffic routes of all six cities. Among these six cities, transportation in Wuhan was influenced by city lockdown to the greatest extent, i.e. by 75.96%. Considering that Wuhan completely forbid public transport within the city [4], this significant reduction of vehicle density proved that the stringent implementation of lockdown policy evidently limited human transmission. Paris and Milan show similar influences on transportation, with reductions of about 65%, while Madrid exhibits a comparatively lower reduction of 47.49% that still approximates to 50%. Finally, London's lockdown policy reduced the vehicle density by 28.66%, while New York City showed the least effect with a reduction of 17.75%.

### C. Transportation Density Change and COVID Stringency Index

It can be seen that the change ratios of vehicle numbers on the arterial roads of these six cities can be ranked as follows: Wuhan>Paris ≈ Milan>Madrid>London>New York (see Table 4). These rankings are quite similar to and in accordance with the countries' COVID stringency index [21] on the RS acquisition days (as shown in Figure 2 (b)), with the city of Wuhan being the only outlier. According to the computation methodology of the COVID stringency index, the index value is determined by a sub-index based on the lockdown policies and their implementation range [21]. Since the stringent Wuhan lockdown limited the spread of the COVID epidemic throughout the whole country of China [3, 6, 17, 56], all of the most stringent policies were implemented in the city of Wuhan, and China's overall COVID stringency index was decreased through the local implementation of stringent policy. A similar situation occurred for the stringency index of New York, which was analyzed separately in the reference [21].

Accordingly, in order to evaluate the correlation between COVID stringency index and transportation density change ratio, we correct these indices to be city-wise, as shown in Figure 6 (a). The policy values were retained and the flag for each sub-index was assigned to be "general" in the city-wise index calculation, since all of these six cities were COVID epicenters in the corresponding countries on those days (the index of New York was used directly, since it is the only city to have a separate index [21]). Using the corrected city-wise COVID stringency index, we conducted a regression analysis between the transportation density change ratios and the indices by means of quadratic polynomial fitting, as shown in Figure 6 (b). It can be observed from these results that the $R^2$ values are both as high as 0.83, which indicates that the reduction in transportation density is highly correlated with the stringency of lockdown policy.

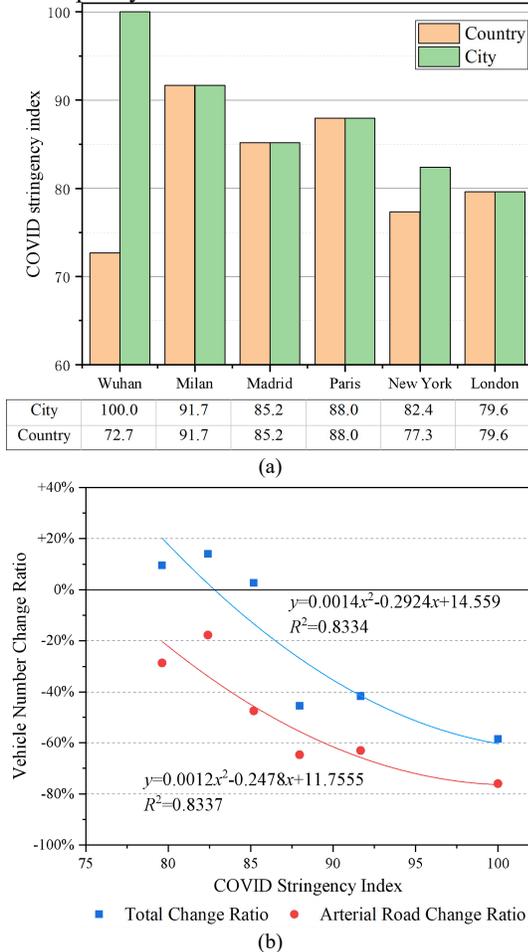

| | Wuhan | Milan | Madrid | Paris | New York | London |
|---|---|---|---|---|---|---|
| City | 100.0 | 91.7 | 85.2 | 88.0 | 82.4 | 79.6 |
| Country | 72.7 | 91.7 | 85.2 | 88.0 | 77.3 | 79.6 |

(a)

(b)

Figure 6: COVID stringency indices for the countries as a whole and the specific cities are shown in (a). The change ratios of vehicle numbers in the six cities and their arterial roads are regressed using the COVID stringency indices in (b).

### D. Distribution of Transportation Density Changes

The statistics of transportation density distribution are presented in Figure 7. From the perspective of the cities' core regions, the transportation density colors become obviously dim in Wuhan, Milan, and Paris; this decline can also be proven with reference to the reduction of vehicle numbers for the whole city in Table 4. In the other three cities, some parts become brighter while others become darker, which can be explained by the fact that more vehicles were stopped in the residential regions.

Examining the transportation density changes, moreover, reveals that in all six cities, the arterial roads exhibit obvious vehicle reduction, since the deep blue mostly appears along the arterial roads. For the three cities that show increasing vehicle numbers in the entire cities (Madrid, New York, and London), the obvious decrease in vehicle numbers is mainly concentrated on the arterial roads, while increases may represent residential regions. For example, Uptown Manhattan in New York City shows a significant increase in vehicle density, which proves that more vehicles were stopped near their homes due to the stay-at-home order; moreover, the center of London's Zone 1 shows vehicle density reduction, while its edges become brighter in the density slice, which may be because there are more apartments in the city center.

## V. DISCUSSION AND CONCLUSION

Our results demonstrate that the lockdown policies during the COVID-19 epidemic clearly reduced transportation density by an average of 49.59%, and as much as 75.96%, by comparing the vehicle numbers before and after lockdown in six large cities (Wuhan, Milan, Madrid, Paris, New York, and London) that were the epicenters of the early global COVID-19 outbreak. The influence on transportation density change ratios can be ranked in size order as follows: Wuhan>Paris≈Milan>Madrid>London>New York; these rankings exhibit extremely high correlations with the COVID lockdown policy stringencies (the $R^2$ is higher than 0.83). We also find that transportation density reductions were spatially relevant to the land-use distribution within the cities.

In order to mitigate the COVID-19 outbreak, similar lockdown policies were implemented in many cities all over the world [4, 7-14, 57]. Studies have indicated that intercity travel bans can delay the progression of the epidemic to other cities by several days [6, 15, 17]. At the same time, control measures within cities, including stay-at-home orders, public transport limitations, closures of schools and workplaces, and bans on public gatherings, have been proven to be effective in decreasing the transmissibility and reducing the case incidence of COVID-19 [6, 18, 20, 58]. One key aspect of these lockdown measures within the city is the control of intracity population transmission [3]. At the moment, most corresponding studies are based on the policy interpretation [6, 15, 17], mathematical modelling [20, 58], and the intercity population flow data [6, 15, 17]; however, there has still been no research on quantitatively evaluating the intracity human transmission affected by city lockdown measures around the world. As we know, control measures during city lockdown will evidently affect the transportation situation [57]. Accordingly, by quantitatively analyzing the transportation density variations in six epicenters of the early COVID-19 epidemic, we find that the transportation densities undergo a significant decrease after city lockdown, and are also strongly positively correlated with the stringency of control measures. Considering that public transport in these cities was largely limited and sometimes even forbidden [4, 40-42], this study indicates that stringent lockdown policy is effective in controlling human transmission. Our study provides new insight into these cities during the epidemic, and can thus help researchers and policy analysts better understand and develop epidemic containment measures to control COVID-19 outbreaks.

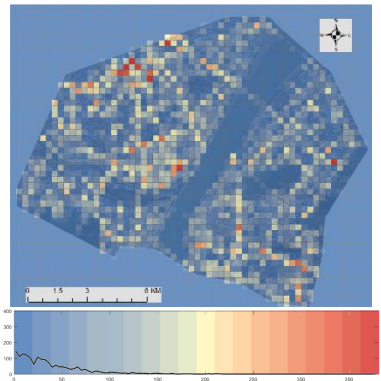

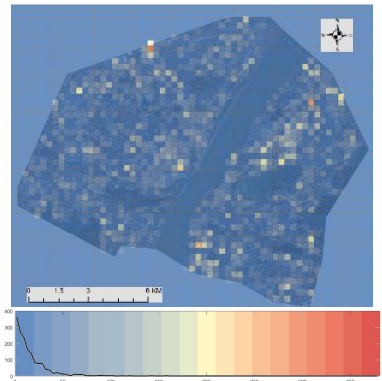

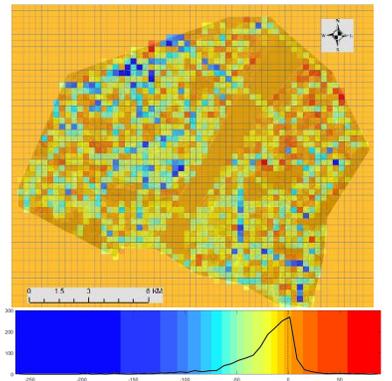

(a)

(b)

(c)

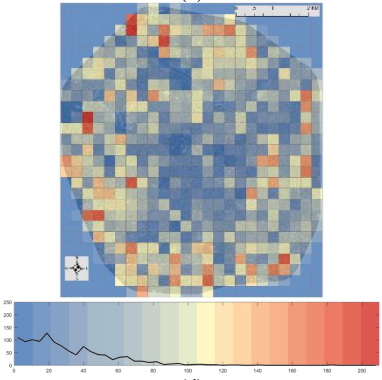

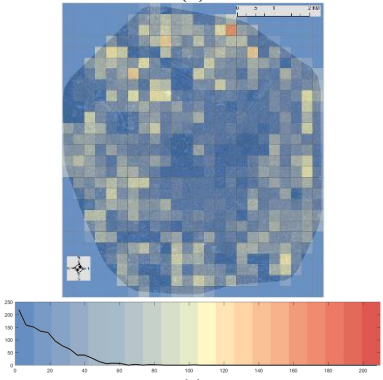

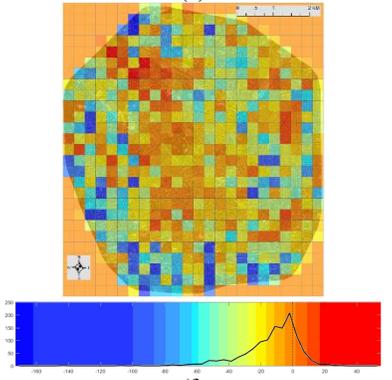

(d)

(e)

(f)

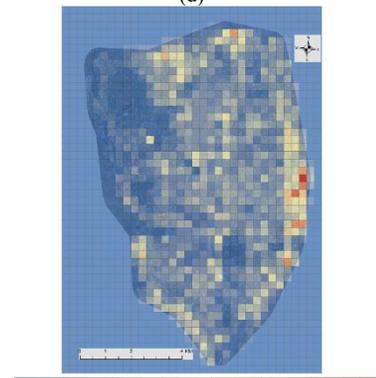

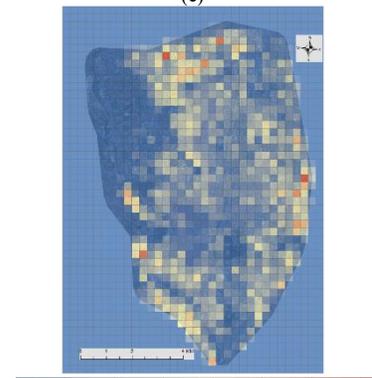

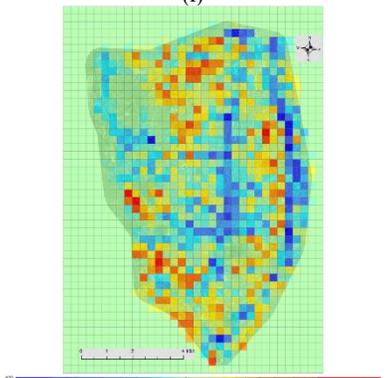

(g)

(h)

(i)

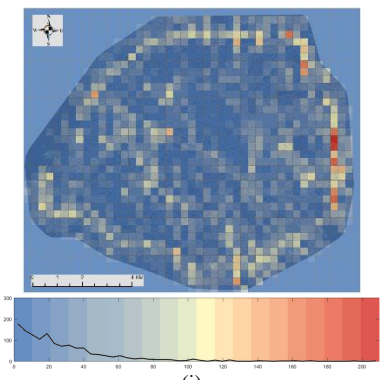

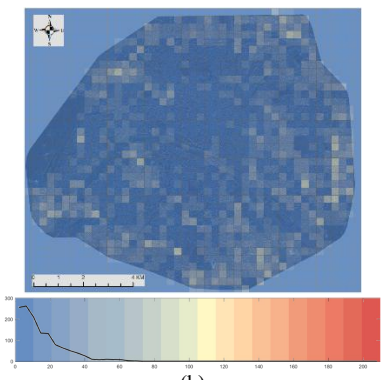

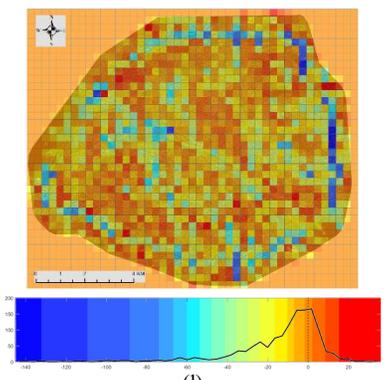

(j)

(k)

(l)

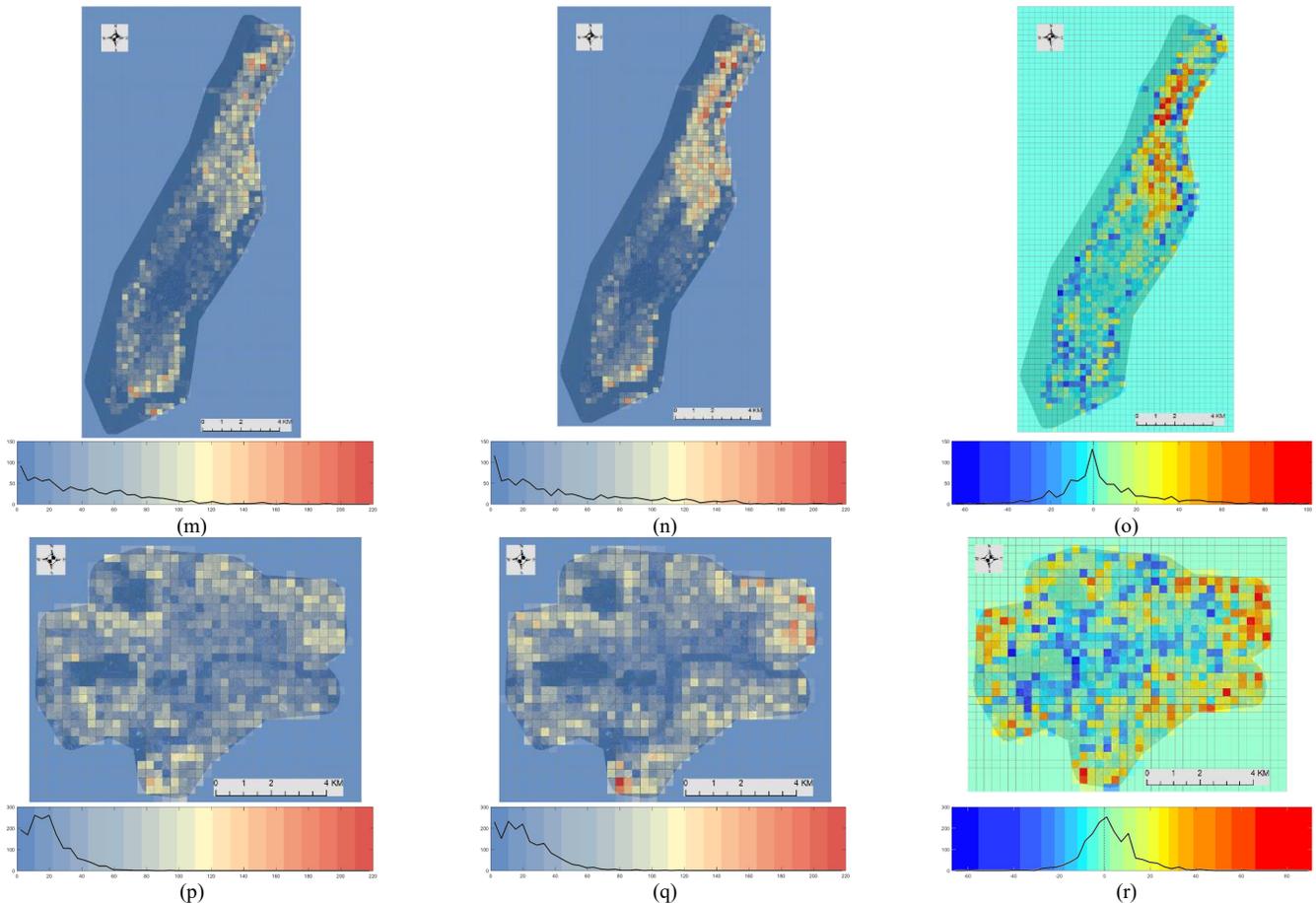

Figure 7: Statistics of vehicle numbers in $300m \times 300m$ blocks in the six cities under analysis. The first column presents the vehicle distributions before lockdown, while the second indicates the vehicle distributions after lockdown. The third column shows the transportation density changes caused by lockdown. The left two columns are presented with equal interval slices, while the last column is shown with Jenks natural break slices for better illustration [59]. The color slices are shown under the transportation density maps; here, the densities in the first two columns of each line are sliced with the same range to facilitate comparison. The rows from top to bottom indicate the statistics for Wuhan, Milan, Madrid, Paris, New York, and London.

Our results may also help in the development of scientific recommendations for lockdown policy implementation. From Figure 6, we may infer that if the lockdown policy is not sufficiently stringent, its effect on controlling transmission will be slight. According to the regression, when the COVID-19 policy stringency is below 74, the numbers of vehicles on the arterial roads may not even reduce. After city lockdown, the transportation density reductions within the cities vary, which may reflect some potential land-use patterns for future city planning.

A recent work also addressed a study on the traffic pattern of cities under lockdown during the COVID-19 epidemic [29]. However, the remote sensing images used in this work is acquired by Planet with the resolution of 3m. For most normal vehicles in cities, they are smaller than 5m×2m, which only occupy less than 2×1 pixels in the image. Besides, this study utilized tophat process to separate vehicles from the roads, which is only sensitive to bright vehicles. So, this previous work may not identify all the vehicles very accurately. It worth noting that this previous work also indicated the obvious reduction of transportation density, and its correlation with lockdown stringency, which shows another evidence for the conclusion of our study.

Our study is also affected by the following limitations. In addition to the detection rate, it is difficult for the automatic vehicle detection algorithm to distinguish between running vehicles on the road and stopped vehicles parked at the roadside. To solve this problem, we also present the statistics of numbers of vehicles on arterial roads, whereas this phenomenon cannot be avoided. Considering that there may be more vehicles parked at the roadside after a city lockdown has been put in place, the transportation density reduction in our study may be underestimated. The second problem is that, for fair comparison, we use OSM data for six cities across the world to extract road regions and define arterial roads. However, OSM data is a free geographic data source that is provided by individuals [39], meaning that we cannot guarantee the accuracy of road type information; this may also affect the fairness when comparing variations in transportation density.

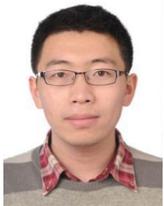

**Chen Wu** (M'16) received B.S. degree in surveying and mapping engineering from Southeast University, Nanjing, China, in 2010, and received the Ph.D. degree in Photogrammetry and Remote Sensing from State Key Lab of Information Engineering in Surveying, Mapping and Remote sensing, Wuhan University, Wuhan, China, in 2015.

He is currently an Associate Professor with the State Key Laboratory of Information Engineering in Surveying, Mapping and Remote Sensing, Wuhan University, Wuhan, China. His research interests include multitemporal remote sensing image change detection and analysis in multispectral and hyperspectral images.

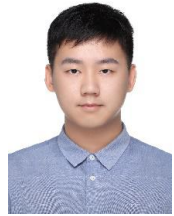

**Sihan Zhu** received the B.S. degree in geographic information science from China University of Petroleum (East China), Qingdao, China in 2019. And he is currently pursuing the M.S. degree in surveying and mapping engineering with the State Key Laboratory of Information Engineering in Surveying, Mapping, and Remote Sensing, Wuhan University, Wuhan, China.

His research interests include remote sensing image processing, transfer learning, and deep learning.

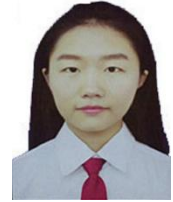

**Jiaqi Yang** (S'20) received the B.S. degree in geographic information science from Dalian Maritime University, Dalian, China, in 2019. She is currently pursuing the M.S. degree at the State Key Laboratory of Information Engineering in Surveying, Mapping, and Remote Sensing, Wuhan University.

Her research interests include deep learning and remote sensing image processing.

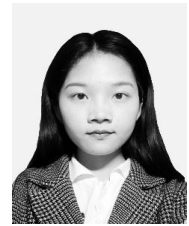

**Meiqi Hu** received the B.S. degree in surveying and mapping engineering from the School of Geoscience and info-physics, Central South University, Changsha, China, in 2019. She is currently pursuing the M.S. degree with the State Key Laboratory of Information Engineering in Surveying, Mapping, and Remote sensing, Wuhan University, Wuhan, China.

Her research interests include deep learning and multitemporal remote sensing image change detection.

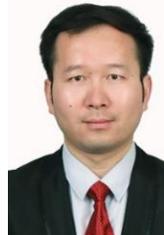

**Bo Du** (M'10–SM'15) received the Ph.D. degree in Photogrammetry and Remote Sensing from State Key Lab of Information Engineering in Surveying, Mapping and Remote Sensing, Wuhan University, Wuhan, China in 2010.

He is currently a professor with the School of Computer Science and Institute of Artificial Intelligence, Wuhan University. He is also the director of and National Engineering Research Center for Multimedia Software Wuhan University, Wuhan, China. He has more than 80 research papers published in the IEEE Transactions on image processing (TIP), IEEE Transactions on cybernetics (TCYB), IEEE Transactions on Pattern Analysis and Machine Intelligence(TPAMI), IEEE Transactions on Geoscience and Remote Sensing (TGRS), IEEE Journal of Selected Topics in Earth Observations and Applied Remote Sensing (JSTARS), and IEEE Geoscience and Remote Sensing Letters (GRSL), etc. Thirteen of them are ESI hot papers or highly cited papers. His major research interests include pattern recognition, hyperspectral image processing, and signal processing.

He is currently a senior member of IEEE. He serves as associate editor for Neural Networks, Pattern Recognition and Neurocomputing. He also serves as a reviewer of 20 Science Citation Index (SCI) magazines including IEEE TPAMI, TCYB, TGRS, TIP, JSTARS, and GRSL. He won the Highly Cited Researcher 2019 by the Web of Science Group. He won the IJCAI (International Joint Conferences on Artificial Intelligence) Distinguished Paper Prize, IEEE Data Fusion Contest Champion, and IEEE Workshop on Hyperspectral Image and Signal Processing Best paper Award, in 2018. He regularly serves as senior PC member of IJCAI and AAAI. He served as area chair for ICPR.

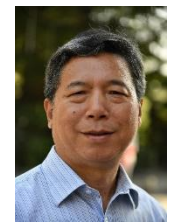

**Liangpei Zhang** (M'06–SM'08–F'19) received the B.S. degree in physics from Hunan Normal University, Changsha, China, in 1982, the M.S. degree in optics from the Xi'an Institute of Optics and Precision Mechanics, Chinese Academy of Sciences, Xi'an, China, in 1988, and the Ph.D. degree in photogrammetry and remote sensing from Wuhan University, Wuhan, China, in 1998.

He is a chair professor in state key laboratory of information engineering in surveying, mapping, and remote sensing (LIESMARS), Wuhan University. He was a principal scientist for the China state key basic research project (2011–2016) appointed by the ministry of national science and technology of China to lead the remote sensing program in China. He has published more than 700 research papers and five books. He is the Institute for Scientific Information (ISI) highly cited author. He is the holder of 30 patents. His research interests include hyperspectral remote sensing, high-resolution remote sensing, image processing, and artificial intelligence.

Dr. Zhang is a Fellow of Institute of Electrical and Electronic Engineers (IEEE) and the Institution of Engineering and Technology (IET). He was a recipient of the 2010 best paper Boeing award, the 2013 best paper ERDAS award from the American society of photogrammetry and remote sensing (ASPRS) and 2016 best paper theoretical innovation award from the international society for optics and photonics (SPIE). His research teams won the top three prizes of the IEEE GRSS 2014 Data Fusion Contest, and his students have been selected as the winners or finalists of the IEEE International Geoscience and Remote Sensing Symposium (IGARSS) student paper contest in recent years. He is the Founding Chair of the IEEE Geoscience and Remote Sensing Society (GRSS) Wuhan Chapter. He also serves as an Associate Editor or Editor for more than ten international journals. He serves as an Associate Editor for the IEEE TRANSACTIONS ON GEOSCIENCE AND REMOTE SENSING.

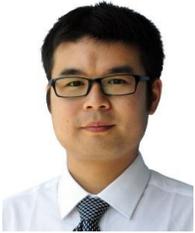

**Lefei Zhang** (S'11-M'14-SM'20) received the B.S. and Ph.D. degrees from Wuhan University, Wuhan, China, in 2008 and 2013, respectively. He is currently a professor with the School of Computer Science, Wuhan University. He was a Big Data Institute Visitor with the Department of Statistical Science, University College London in 2016, and a Hong Kong Scholar with the Department of Computing, The Hong Kong Polytechnic University in 2017. His research interests include pattern recognition, image processing, and remote sensing.

Dr. Zhang serves as an associate editor for Pattern Recognition and IEEE Geoscience and Remote Sensing Letters.

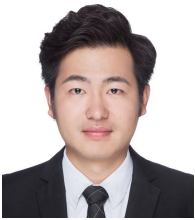

**Chengxi Han** (S'20) received the B.S. degree from the School of Geosciences and Info-Physics, Central South University, Changsha, China, in 2018. He is pursuing the Ph.D. degree in photogrammetry and remote sensing with the State Key Laboratory of Information Engineering in Surveying, Mapping and Remote Sensing, Wuhan University, Wuhan, China.

His research interests include deep learning and remote sensing image processing.

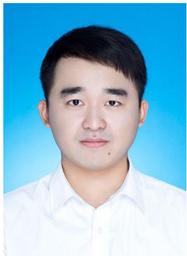

**Meng Lan** (S'19) received the B.S. degree in school of computer science from Wuhan University, Wuhan, China, in 2018. He is now pursuing the Ph.D. degree in school of computer science from Wuhan University, Wuhan, China.

His research interests include deep learning and image processing.